\pdfoutput=1

\documentclass[11pt]{article}

\usepackage{acl}

\usepackage{times}
\usepackage{latexsym}

\usepackage[T1]{fontenc}

\usepackage[utf8]{inputenc}

\usepackage{microtype}

\usepackage{inconsolata}

\usepackage{graphicx}

\usepackage{color, xcolor}

\usepackage{booktabs}
\usepackage{multicol}
\usepackage{multirow, makecell, caption}
\usepackage{colortbl}
\usepackage{tikz}
\usepackage{arydshln}
\usepackage{pgfplots}
\usepackage{amsmath}
\usepackage{amssymb}

\usepackage{tabularx}
\usepackage{enumerate}
\usepackage{tcolorbox}

\usepackage{hyperref}

\makeatletter
\def\adl@drawiv#1#2#3{%
        \hskip.5\tabcolsep
        \xleaders#3{#2.5\@tempdimb #1{1}#2.5\@tempdimb}%
                #2\z@ plus1fil minus1fil\relax
        \hskip.5\tabcolsep}
\newcommand{\cdashlinelr}[1]{%
  \noalign{\vskip\aboverulesep
           \global\let\@dashdrawstore\adl@draw
           \global\let\adl@draw\adl@drawiv}
  \cdashline{#1}
  \noalign{\global\let\adl@draw\@dashdrawstore
           \vskip\belowrulesep}}
\makeatother

\title{Step-by-Step Mastery: Enhancing Soft  Constraint\\ Following Ability of Large Language Models}

\author{
    Qingyu Ren\textsuperscript{1,3}, Jie Zeng\textsuperscript{1}, Qianyu He\textsuperscript{1}, Jiaqing Liang\textsuperscript{2}\thanks{\ Corresponding author.}, Yanghua Xiao\textsuperscript{1}\\\textbf{Weikang Zhou\textsuperscript{3}, Zeye Sun\textsuperscript{3}, Fei Yu\textsuperscript{3}}\\
    \\
    \textsuperscript{1}Shanghai Key Laboratory of Data Science, College of Computer Science and Artificial Intelligence, \\Fudan University 
    \textsuperscript{2}School of Data Science, Fudan University  \textsuperscript{3}Ant Group\\
    \{qyren24, jzeng23, qyhe21\}@m.fudan.edu.cn, \{liangjiaqing, shawyh\}@fudan.edu.cn\\
}

\begin{document}
\maketitle
\begin{abstract}

It is crucial for large language models (LLMs) to  follow  instructions that involve multiple constraints. In real-world scenarios, user instructions often contain soft constraints, which are semantically related and cannot be rule-based verified, posing challenges for LLMs. To  enhance the soft constraint following ability of LLMs, we initially design a pipeline to construct datasets with high-quality outputs for instructions containing soft constraints automatically. Additionally, to fully utilize the positive and negative samples generated during the data construction process, we choose Direct Preference Optimization (DPO) as the training method. Furthermore, taking into account the difficulty of soft constraints indicated by the number of constraints, we design a curriculum learning training paradigm based on the constraint quantity. We experimentally evaluate the effectiveness of our methods in improving LLMs'  soft constraint following ability and analyze the factors driving the improvements. The datasets and code are publicly
available at \href{https://github.com/Rainier-rq/FollowSoftConstraint}{https://github.com/Rainier-rq/FollowSoftConstraint}.

\end{abstract}

\section{Introduction}

In the application of LLMs, the instruction following ability is of paramount importance, especially when the instructions involve multiple constraints~\cite{lou2024large,zeng2023evaluating,zhou2023instruction}. The capability of LLMs plays a critical role in aligning with human preferences, ensuring the reliability and helpfulness of the models' outputs~\cite{wang2023aligning,song2024preference}.

\begin{figure}[t] 
    \centering
        \includegraphics[width=0.5\textwidth]{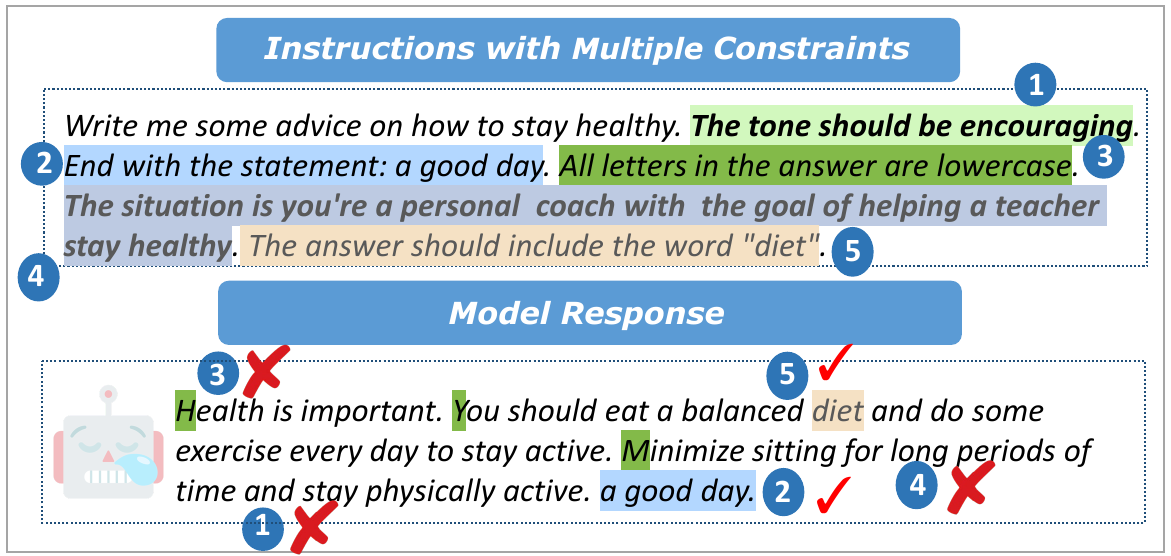}
    \caption{In real-world scenarios, user instructions contain many soft constraints, posing challenges for LLMs. We use \textbf{bold} to represent soft constraints. }
    \label{fig:010intro}
\end{figure}

Following instructions with soft constraints is imperative for LLMs~\cite{jiang2023followbench,qin2024infobench}. Constraints can be categorized into soft and hard constraints. Hard constraints can be explicitly expressed as specific rules and directly verified through programming methods~\cite{zhou2023instruction,he2024complex}. For example, Python can parse JSON data to verify whether it follows specific format constraints.  However, instructions in real-world applications often contain semantic-level limitations, which can be categorized as soft constraints. Soft constraints include restrictions related to content~\cite{liang2024controllable, zhang2023survey}, specific backgrounds~\cite{shanahan2023role, liu2023agentbench}, and the style of expressions~\cite{sigurgeirsson2024controllable, mukherjee2024large}. They are difficult to verify automatically through programming methods. As shown in Fig.~\ref{fig:010intro}, following soft constraints is challenging for LLMs.

\begin{figure*}[t] 
    \centering
            \includegraphics[width=1\textwidth]{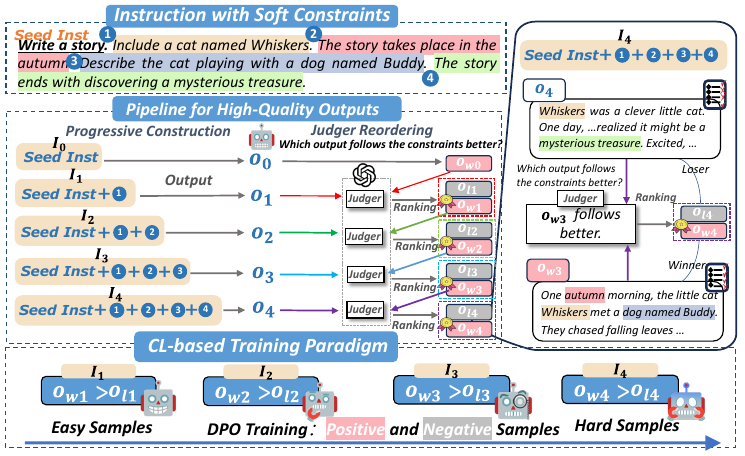}
    \caption{The framework of our study. We first design a pipeline to automatically   construct datasets with high-quality outputs for soft constraint following. Then, we propose a  reinforcement learning training paradigm based on curriculum learning. CL denotes curriculum learning.}
    \label{fig:010method}
\end{figure*}

Existing methods for improving soft constraint following ability of LLMs have following limitations: First, much of the existing work focuses on  evaluating LLMs' soft constraint following ability~\cite{chen2024benchmarking,qin2024infobench} rather than  how to improve the ability. Second, when constructing soft constraint datasets, current methods typically introduce all constraints at once and directly prompt advanced models to generate responses~\cite{xu2023wizardlm,chiang2023vicuna}. However, these responses may not satisfy all  constraints. For example, GPT-4 demonstrates a  constraint satisfaction rate of only 74.4\% on FollowBench~\cite{jiang2023followbench}, making the assurance of high-quality outputs difficult. Third, in terms of the training paradigm, existing work ignores the difficulty of  constraint following indicated by the number of constraints~\cite{he2024complex,qi2024constraint}. Many studies  show that organizing training data in a curriculum-based order—starting from simpler examples and gradually increasing complexity—leads to better performance than random shuffling~\cite{sun2024conifer,lee2024improving}. Therefore, a pipeline constructing  datasets with high-quality outputs and an efficient training paradigm are required.


In this work, we systematically investigate strategies to enhance the ability of LLMs to follow soft constraints, with the framework shown in Fig.~\ref{fig:010method}, including constructing  datasets with high-quality outputs and proposing a reinforcement
learning training paradigm based on curriculum learning. To construct  datasets with  high-quality outputs for soft constraint following, we progressively add constraints to the instructions and incorporate Judger to reorder the outputs based on  constraint following. To fully leverage the positive and negative samples generated during Judger reordering, we apply  Direct Preference Optimization (DPO)~\cite{rafailov2024direct} as the training method. Since soft constraints cannot be verified through rule-based methods, Judger reordering can ensure that the constructed preference dataset adheres to the correct constraint-following behavior. To account for the  difficulty of soft constraints indicated by the number of constraints, we propose a  training paradigm that constructs curricula based on the number of constraints in the instruction. In this paradigm, the model starts by learning simpler tasks (fewer constraints) and  progresses to more complex ones (more constraints). Our methods can improve  LLMs' soft constraint following ability effectively. Moreover, our method is also effective for hard constraints. 

Our contributions are summarized as follows:
(1) We design a pipeline to automatically construct datasets with high-quality outputs for soft constraint following.
(2) We introduce a reinforcement learning training paradigm that constructs curricula based on the number of constraints.
(3) We conduct
extensive experiments to validate the effectiveness
of our methods and analyze the reasons for the performance improvements.

\section{Related Work}
\textbf{Soft Constraint Following} Existing research on soft constraint following  focuses on evaluating the ability of LLMs by constructing benchmarks~\cite{jiang2023followbench,qin2024infobench}. These benchmarks  include a variety of fine-grained constraint types~\cite{zhang2024cfbench}. These constraints can be categorized into several types: (1) Content soft constraints involve restrictions on the scope or depth of the responses~\cite{zhou2023controlled,ashok2024controllable}. (2) Situation soft constraints refer to the  background limitations~\cite{wang2023interactive,shao2023character}. (3) Style soft constraints limit the manner of expressions~\cite{tao2024cat,pu2024style}.  Some works directly utilize responses generated by GPT-4 to construct datasets~\cite{sun2024conifer,peng2023instruction}. Different from these, we propose a pipline  constructing datasets with high-quality outputs to improve LLMs'
soft constraint following ability.

\textbf{Curriculum Learning}
Curriculum learning is a training strategy that mimics the learning process of humans from simpler to more complex tasks~\cite{soviany2022curriculum,wang2021survey}. Current research on LLMs' curriculum learning can be categorized into two primary paradigms: (1) Learning Based on Data Difficulty: This approach organizes the training data sequence according to various evaluation metrics. Metrics such as sequence length~\cite{pouransari2024dataset}, perplexity~\cite{liu2024let} have been employed to guide this process. LLMs can also construct curricula  by organizing the training data sequence in a strategic way~\cite{ryu2024curricullm}. (2) Learning Based on Task Difficulty: This paradigm focuses on changing the training tasks~\cite{chen2024self} or adjusting the training objectives~\cite{zhao2024automatic,lee2024improving}. However, our work organizes the curriculum based on the number of constraints .

\section{Method}\label{sec:method1}
In this section, we provide a detailed explanation of how to obtain datasets with high-quality outputs and how to leverage the dataset by establishing a curriculum learning training paradigm. The framework is shown in Fig.~\ref{fig:010method}.
\subsection{Dataset Construction}\label{sec:method}

We first construct a multi-constraint instruction following dataset. Adding all constraints at once increases the complexity of the instructions rapidly, making it difficult for the model to understand all  constraints~\cite{he2024complex}. To addresss this, we propose a  progressive construction method, adding one constraint at a time, which allows the model to understand and learn to follow each constraint effectively. Moreover, existing works in dataset construction rely on advanced models to directly generate the outputs~\cite{sun2024conifer}. However, even GPT-4 is struggling to follow the instructions with  complex soft constraints~\cite{jiang2023followbench,qin2024infobench}. Since soft constraints cannot be verified for compliance through rules, we  use Judger to reorder the outputs based on the extent of constraint following to obtain high-quality outputs. Overall, our pipeline consists of two successive steps: \textbf{Progressive Construction} and \textbf{Judger Reordering}.

\subsubsection{Progressive Construction }\label{sec:const}

To enable the model to effectively learn how to follow each  constraint, we propose a progressive construction method. Specifically, we add only one constraint at a time,  enabling the model progressively learn to follow each constraint during the training process.

We begin by collecting seed instructions from three sources. We first collect instructions from Open Assistant~\cite{kopf2024openassistant}, which includes instructions generated by users interacting with chatbots. We select rank 0 instructions and those from the first turn of conversations. Next, we gather  manually created instructions from the Self-Instruct~\cite{wang2022self}. The third source is Super-Natural~\cite{wang2022super}, from which we select instructions after filtering out tasks with simple outputs. These three sources together provide a total of 1,500 seed instructions, offering a broad range of coverage across diverse tasks.

Subsequently, we construct different types of soft constraints. Initially, we  categorize the soft constraints into three types: content, situation, and style. Next, we randomly select  constraints for each seed instruction. For the soft constraint type we select, GPT-4o is employed to generate corresponding descriptions. The prompt used to construct soft constraints is detailed in the Appx.~\ref{sec:constraint}. For the hard constraint type, we select the description from a predefined description list.

To obtain multi-constraint instructions, we adopt a  progressive construction approach. As shown in Fig.~\ref{fig:010method}, we add only one constraint to the instruction at a time, allowing the model to focus on learning to follow each  constraint. This  process helps the model gradually adapt to the increasing complexity of the constraint following task and  balance multiple constraints effectively. Specifically, for seed instruction $ I_{0} $, we progressively add one constraint each time to form the instruction set $ I = \{I_1, I_2, \dots, I_n\} $, where 
$n$ denotes the maximum number of constraints. For each instruction $ I_k $ with $ k $ constraints $(k = 1, 2, \dots, n)$, we use GPT-4o to generate the corresponding output $O_k = \text{LLM} (I_k)$. After performing inference on all the instructions in the instruction set 
$ I $, we obtain the output set $ O = \{O_1, O_2, \dots, O_n\} $.

\subsubsection{Judger Reordering}\label{sec:judger}

In \S\ref{sec:const}, we progressively increase the constraints, but the quality of the outputs may not improve incrementally.
To address this, we  introduce Judger to reorder the outputs based on constraint following to ensure the quality of outputs.

During the  progressive construction process in \S\ref{sec:const}, as new constraints are progressively added, the model’s responses may overlook previously added constraints, leading to a decrease in the output quality. As shown in Fig.~\ref{fig:010method}, to obtain high-quality outputs, we introduce Judger, where GPT-4o is prompted to compare two outputs before and after adding the new constraint, to determine which better follows the updated instruction. The two outputs  in each comparison are recorded, and the one deemed better by  Judger is used for the next round of comparison. By iteratively ranking the outputs, the constructed data is consistent with constraint following, thereby improving the output quality.

Specifically, when a new constraint is added into the instruction $ I_{k-1} $ to form $ I_{k} $ , the model's response $O_k$ may not fully follow the constraints in $ I_{k} $. To obtain high-quality outputs, we use Judger to rank the new output $ O_k $ with the previous  output $ O_{w_{k-1}} $ that more follows $ I_{k-1} $ to determine which one better follows the current instruction $ I_k $:
$
O_{w_{k}}, O_{l_{k}} = \text{Judger} \ ( I_k, O_{w_{k-1}}, O_k ).
$

\indent In each ranking, we can obtain the output $ O_{w_{k}}$ which   follows the current instruction $ I_k $ better and the output $ O_{l_{k}}$ which  follows less. Finally, after completing all $ n $ rankings, we obtain the positive set $ O_w = \{O_{w_1}, O_{w_2}, \dots, O_{w_n}\} $, which consists of outputs that  follow their respective instructions better. We also obtain the negative set $ O_l = \{O_{l_1}, O_{l_2}, \dots, O_{l_n}\} $, which contains outputs that less follow. The  prompt  used to reorder outputs and  reordering cases are detailed in the Appx.~\ref{sec:ajudger}.

\subsection{Curriculum Learning Training Paradigm}\label{sec:train}

In  \S\ref{sec:judger}, we use Judger to obtain the positive set $ O_{w} $ and the negative set $ O_{l} $. As shown in Fig.~\ref{fig:010method}, to fully leverage both the positive and negative sets, we apply DPO as the training method.  To account for the  difficulty
of soft constraints indicated by the number of constraints , we establish a curriculum learning training paradigm  based on the number of constraints in the instruction.

Given the positive set and the negative set, we  can construct the  training dataset  with $n$ triplets: ($I_1$, $ O_{w_{1}} $, $ O_{l_{1}}$), ($I_2$, $ O_{w_{2}} $, $ O_{l_{2}}$), \dots, ($I_n$, $ O_{w_{n}} $, $ O_{l_{n}}$). In each triplet, the output from  $ O_{w} $ is preferred than the output from $ O_{l} $. To fully leverage these samples, we apply DPO to train the model. To enable the model to learn  from easy to hard tasks during training, we propose a curriculum learning training paradigm based on the number of constraints, starting with simpler tasks involving fewer soft constraints and progressively advancing to more complex tasks involving more soft constraints. 

Specifically, for curriculum $k$, the training dataset $ D_k $ contains the triplet $ (I_k, O_{w_k}, O_{l_k}) $, where \( I_k \) represents the instruction with $ k $ constraints, and \( O_{w_k} \) and \( O_{l_k} \) denote the corresponding outputs in preference learning. The model begins with the simplest dataset, \( D_1 \), and sequentially progresses through \( D_2 \) to \( D_n \), gradually enhancing its ability to handle more soft constraints.  The complete curriculum is defined as \( D = \{D_1, D_2, \dots, D_n\} \). This stepwise approach ensures that the model first builds a strong foundation by learning from simpler datasets with fewer soft constraints, and gradually adapts to more complex datasets with more constraints. To prevent catastrophic forgetting during training~\cite{mccloskey1989catastrophic}, we mix each curriculum’s data with a proportion of 10k ShareGPT examples based on its data size~\cite{chiang2023vicuna}. Based on the curriculum learning training paradigm, the  loss  function of DPO training is as follows:

\begin{small}
\begin{gather*}
\small
\mathcal{L}_{\text{DPO}}(\pi_\theta;\pi_\text{ref}) = - \tiny{\mathbb{E}}_{(I_k, O_{w_k}, O_{l_k})\sim {D}_{k}}[\text{log}\sigma (\beta \text{log}\frac{\pi_\theta(O_{w_k} | I_k)}{\pi_\text{ref}(O_{w_k} | I_k)} \\
- \beta \text{log}\frac{\pi_\theta(O_{l_k} | I_k)}{\pi_\text{ref}(O_{l_k} | I_k)})].
\end{gather*}
\end{small}

where $\pi_{\theta}$ represents the current model, and $\pi_{\text{ref}}$ denotes the reference model. 

To ensure training stability~\cite{xu2024contrastive}, we add the SFT loss into the DPO loss function:
\begin{equation}
    \mathcal{L}_{\text{Ours}} = \mathcal{L}_{\text{DPO}} + \mathcal{L}_{\text{SFT}}.\nonumber
\end{equation}

\indent where SFT loss is as follows:
\begin{equation}
    \mathcal{L}_{\text{SFT}}(\pi_{\theta}) = - \mathbb{E}_{(I_k, O_{w_k}) \sim D_k}[\log \pi_{\theta}(O_{w_k} | I_k)].\nonumber
\end{equation}

\subsection{Analysis and Comparison}
\newcolumntype{g}{>{\columncolor{green!10}}c}
\setlength\tabcolsep{7pt}
\begin{table}[t]
\centering
\huge
\newcolumntype{b}{>{\columncolor{blue!10}}c}
\renewcommand{\arraystretch}{1.1}
\resizebox{0.5\textwidth}{!}{
\begin{tabular}{cccccc}
\toprule
 Curriculum      &\# Constraints         & \# Preference Pairs & Avg Length  \\ \midrule
Curri.1                   &   1        &3714&369\\
Curri.2          &    2     &3494 &422      \\ 
Curri.3  &3     &3387& 461   \\
Curri.4           &  4      &3300   & 503 \\
Curri.5              &  5     &3148 & 516    \\ 
\bottomrule
\end{tabular}
}
 \caption{Statistics of curricula.  \# Constraints refers to the number of constraints in each instruction. \# Preference Pairs refers to the number of preference pairs. “Avg Length” denotes the
average instruction length. 
}
 \label{fig:num}
\end{table}
\begin{figure}[t]
\centering
\includegraphics[width=1.0\columnwidth]{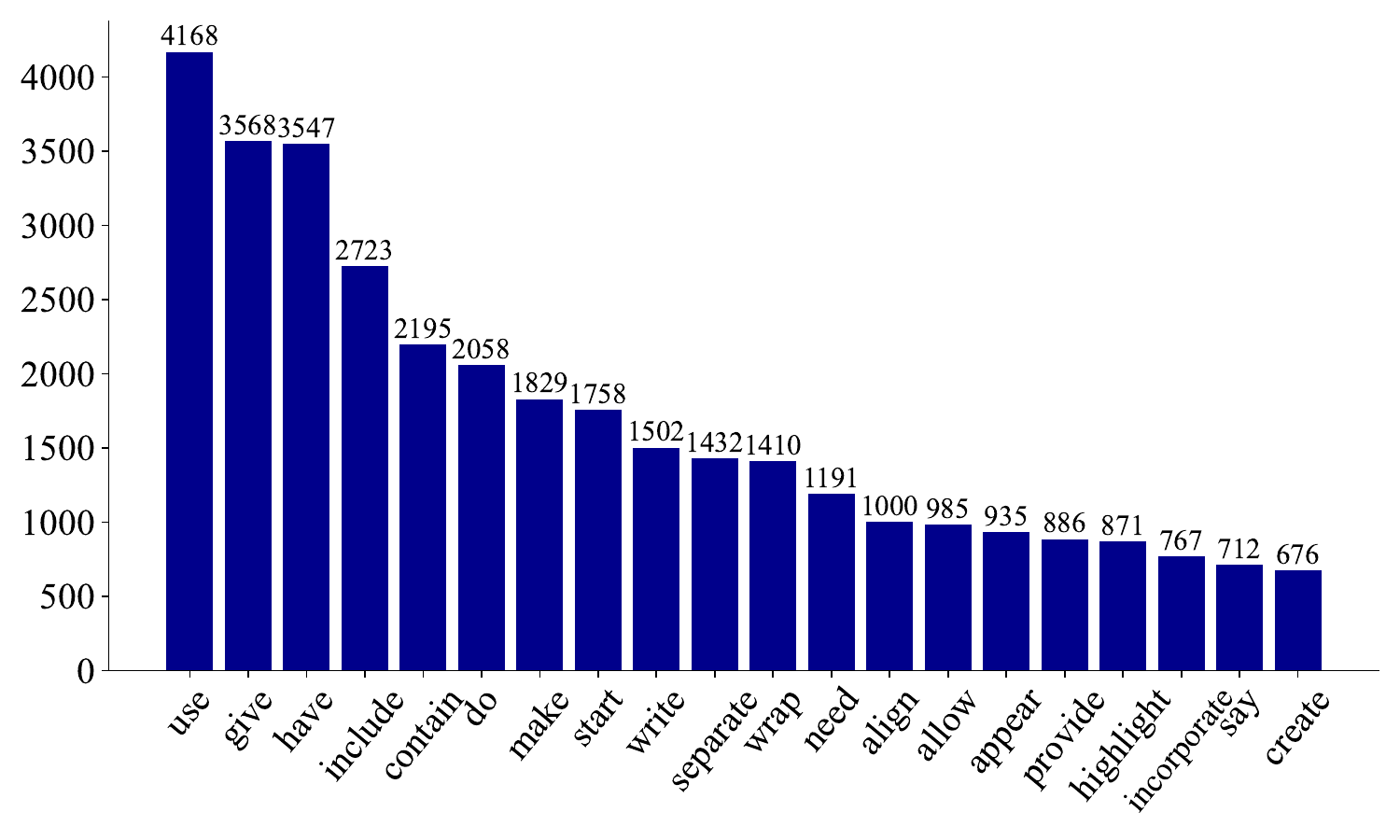}
 \caption{The verb frequency in the instructions.}
 \label{fig:diver}
\end{figure}

\subsubsection{Data Statistics}
We present a statistical analysis of different curricula in Tab.~\ref{fig:num}. The results  show that the number of constraints and instruction length in the curriculum continuously increase. Each curriculum contains a large scale of preference data. To show the diversity of our dataset, we analyze the frequency of verbs in the instructions. As shown in Fig.~\ref{fig:diver},  the instructions contain a variety of verbs, reflecting diverse linguistic patterns. This diversity is crucial for enhancing the model's ability to generalize across different types of constraints. 






\newcolumntype{g}{>{\columncolor{green!6}}c}
\setlength\tabcolsep{7pt}
\begin{table}[t]
\centering
\huge
\newcolumntype{b}{>{\columncolor{blue!10}}c}
\renewcommand{\arraystretch}{1.2}
\resizebox{0.5\textwidth}{!}{

\begin{tabular}{lccccc}

\toprule
\multicolumn{1}{c}{Method} &                        Nums.       & Cons.     & Reor.  & Prog.                          \\ \midrule
Conifer~\cite{sun2024conifer}                                       & 13606      &  H/S      & $ \times $     & $ \checkmark $                      \\
 Suri~\cite{pham2024suri}                                  & 20000       & S      &  $ \times $     & $ \times $                       \\
AutoIF~\cite{dong2024self}                                & -      & H    & $ \times $      &     $ \times $                   \\

Complex to Simple~\cite{he2024complex}                      & 12939      &  H  & $ \times $   & $ \times $                    \\ 
\rowcolor{green!10} Ours                            & 17043       & H/S      & $ \checkmark $     & $ \checkmark $ \\
\bottomrule
\end{tabular}%
}
\caption{
  A detailed comparison of related works. Ours
represents our dataset.  “Nums.”, “Cons.”, “Reor.”,
and “Prog.”\  denote the number of preference pairs, 
constraint types, whether to perform output reordering, and whether the dataset is progressively constructed.
}

  \label{tab:sta}
\end{table}

\subsubsection{Comparison with Other Works}
As shown in Tab.~\ref{tab:sta}, we compare our dataset with related works. Our dataset is large in scale compared to these methods. In terms of constraint categories, it includes both soft and hard constraints,  enhancing the model's ability to learn to follow different types of constraints. Additionally, we use Judger for pairwise comparisons of outputs, improving the overall quality of the dataset. Moreover, our dataset is progressively constructed. 


\section{Experiments}

\begin{table*}[t]
\newcolumntype{g}{>{\columncolor{green!10}}c}

\newcolumntype{b}{>{\columncolor{blue!10}}c}
\renewcommand{\arraystretch}{0.75} 
\renewcommand{\familydefault}{\rmdefault}
\resizebox{\textwidth}{!}{
\begin{tabular}{lcccccbccccg}

\toprule
\multicolumn{1}{c}{\multirow{2}{*}{Model}} & \multicolumn{6}{c}{FollowBench (HSR)} & \multicolumn{5}{c}{IFEval} \\ \cline{2-12} 
\multicolumn{1}{c}{} & L1 & L2 & L3 & L4 & L5 & Avg & \text{[S]P} & \text{[S]I} & \text{[L]P} & \text{[L]I} & Avg \\ \midrule
GPT-4~\cite{achiam2023gpt}$^{*}$                                                              & 84.7  & 75.6  & 70.8  & 73.9  & 61.9  & 73.4  & 76.9     & 83.6     & 79.3     & 85.4     & 81.3 \\

GPT-3.5 Turbo$^{*} $                                  & 80.3  & 68.0  & 68.6  & 61.1  & 53.2  & 66.2  & -     & -     & -     & -     & - \\ \cdashlinelr{1-12}
WizardLM-v1.2-13B~\cite{xu2023wizardlm} & 56.4 &49.2& 37.0& 33.1& 24.2& 40.0 & 43.6 & 54.4 & 48.4 & 59.1 & 51.4\\
Vicuna-13B-v1.5~\cite{chiang2023vicuna} & 56.2 & 42.9 & 32.3 & 32.1 & 24.6 & 37.6 & 43.1 & 53.6 & 46.6 & 58.0 & 50.3 \\
Conifer-7B~\cite{sun2024conifer}       &   54.3 &49.5 &49.3& 40.8& 30.5 &44.9     &   45.8& 57.1 &50.8& 62.0 &53.9\\
Conifer-7B-DPO~\cite{sun2024conifer} & 60.3 &53.6& 48.0& 47.1& 41.0& 50.0& 48.1 & 59.1&52.3& 63.3&55.7
\\
Mistral\textsubscript{CRAB}+DPO~\cite{qi2024constraint}&
 66.1 &53.6& 53.4& 42.4& 31.7& 49.4& 49.7& 61.5& 57.7& 68.5& 59.3
\\
Mistral-7B-ShareGPT~\cite{jiang2023mistral} &51.6& 45.8& 38.3& 25.8& 20.7& 36.5& 43.6& 53.5 &47.3& 57.8& 50.6\\
Llama3\textsubscript{CRAB}+DPO~\cite{qi2024constraint}& 64.6& 49.0& 41.6& 35.8& 36.8& 45.5& 40.3& 52.0& 47.7& 58.9& 49.7
\\
\cdashlinelr{1-12}
Mistral-7B-Instruct-v0.2\textsubscript{BASE} & 58.1
&52.1&44.6&39.8&29.9&44.9&44.5&56.4&49.2&61.5&52.9 \\
Mistral-7B-Instruct-v0.2\textsubscript{SFT+Judger} & 58.1&50.5&44.6&36.0&36.2&45.1&44.2&56.5&47.5&60.2&52.1 \\
Mistral-7B-Instruct-v0.2\textsubscript{DPO+Judger+CL} & 58.8&54.8&46.3&39.4&35.0&46.9&46.6&58.5&52.1&63.7&55.2 \\
\cdashlinelr{1-12}
Mistral-7B-Instruct-v0.3\textsubscript{BASE} & 61.0 & 49.3 & 49.0 & 39.7 & 35.0 & 46.8 & 47.0 & 58.0 & 52.1 & 62.7 & 55.0 \\
Mistral-7B-Instruct-v0.3\textsubscript{SFT+Judger} & 58.7 & 52.4 & 42.5 & 37.2 & 35.6 & 45.3 & 56.8 & 67.8 & 60.6 & 71.3 & 64.1 \\
Mistral-7B-Instruct-v0.3\textsubscript{DPO+Judger+CL} & 63.3 & 56.0 & 47.5 & 40.0 & 36.7 & 48.7 & 53.4 & 63.5 & 57.1 & 67.5 & 60.4 \\
\cdashlinelr{1-12}
LLaMA3-8B-Instruct\textsubscript{BASE} & \underline{67.8} & 54.5 & 46.6 & \textbf{50.6} & 39.1 & 51.7 & 67.5 & 76.1 & 72.8 & \underline{80.9} & 74.3 \\
LLaMA3-8B-Instruct\textsubscript{SFT+Judger} & 66.3 & \underline{55.4} & \underline{50.1} & \underline{49.7} & \underline{39.8} & \underline{52.3} & \underline{70.4} & \underline{77.8} & \underline{73.2} & 80.1 & \underline{75.4} \\
LLaMA3-8B-Instruct\textsubscript{DPO+Judger+CL} & \textbf{69.2} & \textbf{59.6} & \textbf{50.8} & 48.9 & \textbf{44.6} & \textbf{54.6} & \textbf{72.5} & \textbf{80.3} & \textbf{77.1} & \textbf{84.1} & \textbf{78.5} \\

\bottomrule

\end{tabular}
}
\caption{
The overall performance on FollowBench and IFEval. We use
\textbf{bold} for the best results and \underline{underlined} for the second-best results among the models ranging from 7B to 13B parameter sizes.
}
\label{tab:main}
\end{table*}

We conduct extensive experiments to validate the effectiveness of our method on improving LLMs' soft constraint following ability.

\subsection{Experiment Setup}

\textbf{Models.} 
We conduct experiments on several widely recognized base LLMs, including the LLaMA series~\cite{dubey2024llama}, Mistral series~\cite{jiang2023mistral}, and Qwen2.5 series~\cite{yang2024qwen2}. Within our experimental framework, we compare three approaches: (1) \textbf{BASE} directly utilizes the base model to generate outputs. (2) \textbf{SFT+Judger} applies supervised fine-tuning on LLMs using the instruction-response pairs $(I_{n},O_{w_n})$ generated by Judger (\S\ref{sec:const},\S\ref{sec:judger}). (3) \textbf{DPO+Judger+CL} utilizes Judger to obtain high-quality training data, which is then used for DPO training following the curriculum learning training paradigm (\S\ref{sec:judger}, \S\ref{sec:train}). For baseline comparisons, we include proprietary models such as GPT-4~\cite{achiam2023gpt} and GPT-3.5 Turbo. Additionally, we also select models specifically designed to enhance the ability to follow multi-constraint instructions.

\noindent \paragraph*{Settings.}  For each seed instruction, we progressively add five constraints. In the curriculum setting, we combine curriculum 1 to 3 into a simple curriculum and curriculum 4 to 5 into a difficult curriculum. The model is first trained on the simple curriculum, and then on the difficult one.

\noindent \paragraph*{Evaluation Benchmarks.}
FollowBench~\cite{jiang2023followbench} is a benchmark that evaluates the ability of models to follow both soft and hard constraints across multiple levels of granularity. CFBench~\cite{zhang2024cfbench} is a benchmark that evaluates the ability of models to follow soft constraints. Each example needs to be evaluated by GPT-4. IFEval~\cite{zhou2023instruction} is a benchmark designed to evaluate the ability to follow hard constraints. 

\begin{figure}[t] 
    \centering
    \includegraphics[width=\columnwidth,height=0.6\columnwidth]{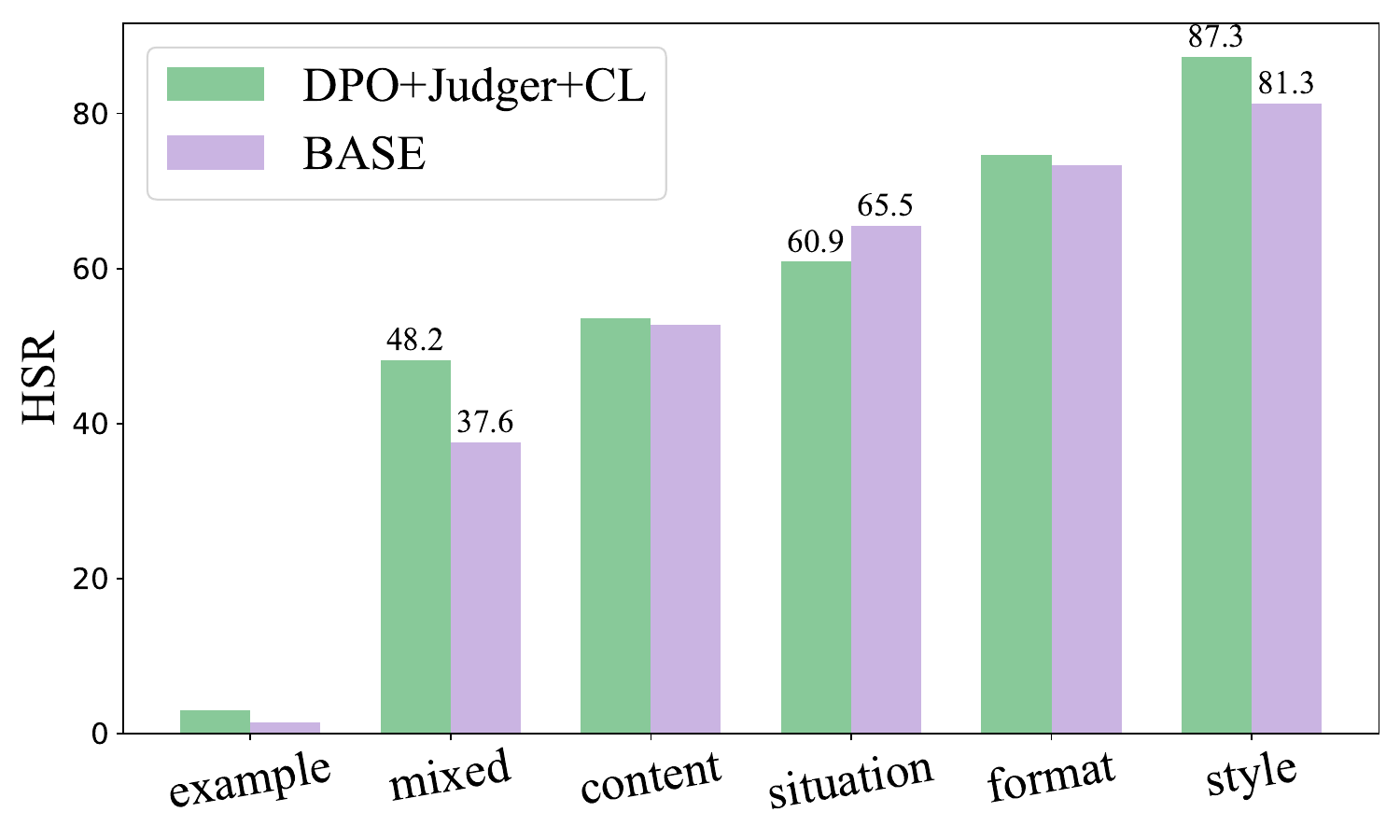} 
    \caption{Results across various constraint categories.}
    \label{fig:010bar}
\end{figure}

\begin{table*}[t]
\newcolumntype{g}{>{\columncolor{green!10}}c}

\newcolumntype{b}{>{\columncolor{blue!10}}c}
\renewcommand{\arraystretch}{0.75} 
\renewcommand{\familydefault}{\rmdefault}
\resizebox{\textwidth}{!}{
\begin{tabular}{lcccccbcccccg}

\toprule
\multicolumn{1}{c}{\multirow{2}{*}{Model}} & \multicolumn{6}{c}{Hard  Satisfaction Rate}         & \multicolumn{6}{c}{Soft  Satisfaction Rate}                       \\ \cline{2-13} 
\multicolumn{1}{c}{}                       & L1    & L2    & L3    & L4    & L5    & Avg    & L1    & L2    & L3    & L4    & L5    & Avg    \\ \midrule

Mistral-7B-Instruct-v0.2\textsubscript{BASE} & 70.5& 58.4&50.6&51.7&39.0&54.0 & 70.5& 67.5 & 66.1 & 64.1 & 61.2 & 65.9 \\
Mistral-7B-Instruct-v0.2\textsubscript{DPO+Judger+CL} & 70.3 & 65.4 & 55.8 & 47.4 & 41.5 & 56.1 & 70.3 & 71.3 & 67.1 & 64.1 & 63.9 & 67.3 \\
\cdashlinelr{1-13}
LLaMA3-8B-Instruct\textsubscript{BASE} & 81.9 & 60.5 & 60.7 & 55.4 & 52.0 & 62.1 & 81.9 & 66.4 & 69.5 & 68.7 & 69.3 & 71.2 \\
LLaMA3-8B-Instruct\textsubscript{DPO+Judger+CL} & 81.2 & 67.5 & 64.9 & 56.3 & 49.9 & 64.0 & 81.2 & 73.3 & 73.3 & 70.3 & 69.3 & 73.5 \\
\cdashlinelr{1-13}
Qwen2.5-32B-Instruct\textsubscript{BASE} & \underline{89.6}&	\underline{78.5}&	\textbf{79.6}&	\underline{70.3}&	\underline{63.7}&	\underline{76.3} & \underline{89.6} & \underline{85.0} & \underline{82.0} & \textbf{82.0} &  \underline{75.1} & \underline{82.7} \\
Qwen2.5-32B-Instruct\textsubscript{DPO+Judger+CL} & \textbf{91.2} & \textbf{82.6} & \underline{78.6} & \textbf{72.2}& \textbf{69.0} & \textbf{78.7} & \textbf{91.2} & \textbf{85.1} & \textbf{84.1} & \underline{79.5} & \textbf{80.4} & \textbf{84.1} \\
\bottomrule

\end{tabular}
}
\caption{
The overall performance on the soft constraint subset of FollowBench. We use
\textbf{bold} for the best results and \underline{underlined} for the second-best results.
}
\label{tab:main_soft}
\end{table*}
\begin{table*}[t]
\newcolumntype{g}{>{\columncolor{green!10}}c}
\newcolumntype{b}{>{\columncolor{blue!10}}c}
\renewcommand{\arraystretch}{0.85}
\renewcommand{\familydefault}{\rmdefault}
\resizebox{\textwidth}{!}{
\begin{tabular}{lcccccccccc}
\toprule
\multirow{2}{*}{Models} & \multicolumn{3}{c}{Easy Set} & \multicolumn{3}{c}{Hard Set} & \multicolumn{3}{c}{Full Set} & \multirow{2}{*}{Avg.} \\
\cmidrule(lr){2-4} \cmidrule(lr){5-7} \cmidrule(lr){8-10}
& CSR & ISR & PSR & CSR & ISR & PSR & CSR & ISR & PSR & \\
\midrule

\rowcolor{blue!10}
BASE & 0.770 & 0.450 & 0.530 & 0.640 & 0.160 & \textbf{0.320} & 0.700 & 0.300 & 0.430 & 0.478 \\
\rowcolor{green!10}
DPO+judger+CL & \textbf{0.790} & \textbf{0.480} & \textbf{0.560} & \textbf{0.670} & \textbf{0.190} & 0.320 & \textbf{0.730} & \textbf{0.330} & \textbf{0.440} & \textbf{0.501} \\
\bottomrule
\end{tabular}
}
\caption{
The overall performance of LLaMA3-8B-Instruct on CFBench. We use \textbf{bold} for the best results.
}
\label{tab:cfbench}
\end{table*}
\subsection{Main Results}

Our method (DPO+Judger+CL) significantly enhances the model's ability to follow soft constraints across different models. Moreover, our method is also effective
for hard constraints. As shown in Tab.~\ref{tab:main}, on FollowBench, which includes both soft and hard constraints, our method improves the model's performance. The  improvement is significant on difficult constraint following tasks, especially at the L5 difficulty level in FollowBench. Specifically, LLaMA-3-8B-Instruct shows an improvement of 5.5\% at the L5 difficulty level. For soft constraints, as shown in Tab.~\ref{tab:main_soft} and Tab.~\ref{tab:cfbench}, we validate the effectiveness of our method using the soft constraint subset of FollowBench and CFBench. The results show that our method is also effective in improving performance on soft constraints. For hard constraints, model's performance on IFEval which  includes only hard constraints demonstrates the effectiveness of our method. After training with  SFT+Judger, there may be a decline in the performance. This drop is attributed to the model's integration of various specialized training techniques during its initial training phase.

From the perspective of constraint types, as shown in Fig.~\ref{fig:010bar}, our method significantly improves the model's performance across different types of constraints. The most notable improvement is observed in the Mixed category, which is defined as a composition of multiple constraint categories to simulate real-world scenarios. Our method enhances the model's performance by 10.6\% in this category, suggesting a notable enhancement in the model's ability to handle complex constraints in real-world scenarios. Moreover, our method improves  model's performance by 6\% in the Style category which contains only soft constraints. In the category of situation constraints, the decrease in performance is due to the fact that our constructed situation constraints mainly focus on soft constraints. As a result, our method struggles with hard constraints in this category, such as complex situational reasoning.

\subsection{Generalization Experiments}
\newcolumntype{g}{>{\columncolor{green!10}}c}
\setlength\tabcolsep{7pt}
\begin{table}[t]
\centering
\huge
\newcolumntype{b}{>{\columncolor{blue!10}}c}
\renewcommand{\arraystretch}{1.25}
\renewcommand{\familydefault}{\rmdefault}
\resizebox{0.5\textwidth}{!}{
\begin{tabular}{lcbg}

\toprule 

\multicolumn{1}{l}{Model} & \multicolumn{1}{l}{BaseModel}&\multicolumn{1}{l}{AlpacaEval2.0}&\multicolumn{1}{l}{MT-Bench}\\

\midrule
GPT-4-0613$^{*}$                   & GPT       & 30.2  &9.18      \\
GPT-3.5-Turbo-0613$^{*}$           & GPT       & 22.4   &8.39     \\ \cdashlinelr{1-4}
LLaMA-3.1-70B-Instruct$^{*}$ & LLaMA3    & 39.3  &8.22     \\
WizardLM-13B-v1.2$^{*}$            & LLaMA2    & 14.5   &7.20      \\
Vicuna-13B-v1.5$^{*}$              & LLaMA2    & 10.5    &6.57   \\ \cdashlinelr{1-4}
BASE          & LLaMA3    & 21.6   &  6.78  \\
DPO+Judger+CL         & LLaMA3    & 22.0   & 6.80  \\ \bottomrule
\end{tabular}
}
\caption{
  Results of the length control win rate of AlpacaEval2.0 and the score of MT-Bench. $^{*}$ indicates that the
results are directly sourced from the original leaderboards.
  }
  \label{tab:general}
\end{table}
Besides the ability to follow soft constraints, we also assess the model's general instruction following abilities on AlpacaEval2.0~\cite{zhao2024long} and MT-Bench~\cite{zheng2023judging}. We first perform SFT on LLaMA3-8B-Instruct, followed by DPO+Judger+CL. Specifically, we use precomputed outputs of GPT-4 Turbo on AlpacaEval as reference outputs and GPT-4o as evaluators. As shown in the Tab.~\ref{tab:general}, our method improves the model's general instruction following ability on both banchmarks.

\subsection{Ablation Studies}
\newcolumntype{g}{>{\columncolor{green!10}}c}

\newcolumntype{b}{>{\columncolor{blue!10}}c}
\renewcommand{\arraystretch}{1}
\begin{table}[t]
\renewcommand{\arraystretch}{1.1} 
\renewcommand{\familydefault}{\rmdefault}
\resizebox{1.0\columnwidth}{!}{%
\begin{tabular}{lccbg}

\toprule
\multicolumn{1}{c}{\multirow{2}{*}{Model}} & \multicolumn{3}{c}{FollowBench (HSR)} & \multicolumn{1}{c}{IFEval} \\ \cline{2-5} 
\multicolumn{1}{c}{}                       & L1 - L3       & L4 - L5      & Avg        & Avg                        \\ \midrule
BASE                                       & 56.3       & 44.9      & 51.7      & 74.3                      \\
 SFT                                  & 59.5       & 38.4      & 51.0      & 73.8                     \\
SFT+Judger                                & 57.3       & 44.8      & 52.3      & 75.4                      \\
DPO+Judger                             & 58.8       & 44.6      & 53.1      & 78.4                     \\
DPO+Judger+CL                      & 59.9       &  46.8  & \textbf{54.6}    & \textbf{78.5 }                    \\ \bottomrule
\end{tabular}%
}
\caption{
  Ablation study results on FollowBench and IFEval.
  }
  \label{tab:abla}
\end{table}

In this section, we conduct ablation experiments to study the impact of Judger and curriculum learning on LLaMA3-8B-Instruct. As shown in Tab.~\ref{tab:abla}, directly using the constructed data  for SFT without Judger reordering underperforms  the SFT+Judger method on both benchmarks, even worse than the base model. The performance decreases significantly at the L4-L5 levels of FollowBench. This  suggests that Judger  plays a critical role in ranking the model's responses to  challenging instructions.
The model trained with DPO outperforms the SFT baseline, especially on IFEval, emphasizing the effectiveness of  DPO  over SFT in constraint following tasks. Additionally, the curriculum learning training paradigm improves the model’s ability to follow constraints on both benchmarks, particularly those at higher difficulty levels (L4-L5)  of FollowBench. This validates the necessity of  curriculum learning paradigm for enhancing the model’s ability to follow soft constraints.

\newcolumntype{g}{>{\columncolor{green!10}}c}
\newcolumntype{b}{>{\columncolor{blue!10}}c} 
\begin{table}[t]
\renewcommand{\arraystretch}{1.2} 
\renewcommand{\familydefault}{\rmdefault}
\resizebox{1.0\columnwidth}{!}{%
\begin{tabular}{lccccccc} 

\toprule
\multicolumn{1}{c}{\multirow{2}{*}{Method}} & \multicolumn{6}{c}{FollowBench (HSR)} &  \\ \cline{2-8} 
\multicolumn{1}{c}{}                       & L1       & L2 & L3 & L4 & L5   & \cellcolor{blue!10}Avg                            \\ \midrule
BASE                   &          67.8& 54.5 & 46.6 & 50.6 & 39.1  & \cellcolor{blue!10} 51.7  \\
w/o  progressive                            & 69.8     & 56.7 & 50.8 & 42.4 & 47.6 & \cellcolor{blue!10} 53.5      \\ 
w/ progressive                              & 69.2     & 59.6 & 50.8 & 48.9 & 44.6 & \cellcolor{blue!10} \textbf{54.6}      \\ 

\bottomrule
\end{tabular}%
}
\caption{
  Results of different  construction methods.
  }
  \label{tab:prog}
\end{table}

\subsection{Analysis}

\subsubsection{The Role of Progressive Construction}
In this section, we analyze the role of progressively constructing the dataset. We compare two methods: w/o progressive and w/ progressive. w/o progressive means replacing both the instruction field and the chosen field with the corresponding ones from samples that include all constraints in the progressively constructed DPO dataset. As shown in Tab.~\ref{tab:prog}, our progressive construction method enhances the model's performance on FollowBench, especially at the difficlut level L4. Introducing all constraints at once  makes the training process unstable. By progressively adding constraints, the model can focus on learning to follow one constraint at a time and learn how to balance  constraints effectively.

\newcolumntype{g}{>{\columncolor{green!10}}c}
\setlength\tabcolsep{7pt}

\newcolumntype{b}{>{\columncolor{blue!10}}c}
\renewcommand{\arraystretch}{1.4}

\begin{table}[t]
\renewcommand{\arraystretch}{1.1}
\centering
\resizebox{1.0\columnwidth}{!}{%
\begin{tabular}{lbg}
\toprule
\multicolumn{1}{l}{Ranking Method} & \multicolumn{1}{l}{Kendall Tau Coefficient}&\multicolumn{1}{l}{Position Consistency} \\ \midrule
 w/o Judger                         & 0.847        &0.743                            \\ 
w/ Judger                          & \textbf{0.862}             &\textbf{0.794}                     \\ \bottomrule
\end{tabular}
}
\caption{
  Results on Judger's effectiveness.
  }
  \label{tab:eval_j}
\end{table}

\begin{table}[ht]
\centering
\small
\begin{tabular}{p{1.6cm}p{5cm}}
\toprule
\multicolumn{2}{c}{\textbf{Judger Ranking Case}} \\  
\midrule
\textbf{Instruction} & 
Please write a description of quantum physics, including "wave function" and "superposition state", avoiding complex formulas, under 100 words. Ensure this description is easy to understand.\\
\midrule
\multirow{2}{*}{\textbf{Responses}} & 
(a) Quantum physics studies the behavior of tiny particles. The *wave function* describes a particle’s state, while *superposition state* means a particle can exist in multiple states at once. \\
& 
(b) Quantum physics studies the behavior of tiny particles. In the quantum world, particles don’t have fixed positions or states; they can be in many states at once, and only take a definite state when measured. \\
\midrule
\textbf{Judger} &
Output (a) is better. It includes required terms, explains them simply, and follows all constraints. Output (b) misses "wave function" and is less precise. \\
\bottomrule
\end{tabular}
\caption{The case for Judger ranking outputs.}
\label{tab:cases}
\end{table}
\begin{figure}[t]
\includegraphics[width=\columnwidth, height=0.3\columnwidth]{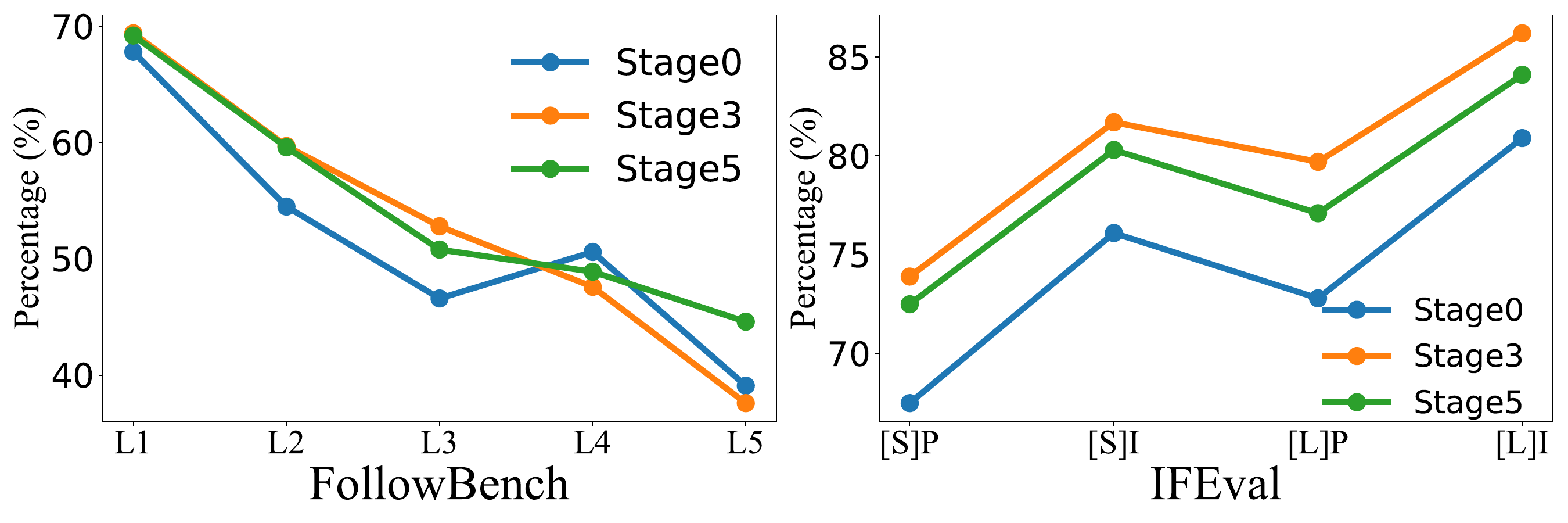}
 \caption{Results of the model  across different training stages in curriculum learning.}
 \label{fig:foll}
\end{figure}

\subsubsection{The Role of Judger}
In this section, we investigate the role of Judger in obtaining high-quality outputs. Specifically, we randomly select 100 seed instructions and rank the outputs at each step of  progressive construction. We evaluate the rankings in three  scenarios: (1) w/o Judger, (2) w/ Judger and (3) rankings annotated by human experts, which serve as the reference standards. As shown in Tab.~\ref{tab:cases}, Output(a) represents the model's initial response generated without the newly introduced constraint “Ensure that this description is easy to understand” while Output(b) shows the response produced after incorporating this additional constraint. In the w/o Judger setting, it assumes Output(b) to be superior by default, without performing reordering. To assess the similarity between the rankings, we employ two metrics. The first is the \textbf{Kendall Tau Coefficient}~\cite{kendall1938new}, which measures the correlation between two rankings by assessing the agreement in the order of paired items. Formally, given two rankings $R_1$ and $R_2$ over the same set of $n$ items. Kendall Tau Coefficient is defined as


\vspace{0.6em} 
\resizebox{0.9\columnwidth}{!}{%
$\displaystyle
\tau(R_1,R_2)
=
\frac{\displaystyle\sum_{1 \le i < j \le n}
\operatorname{sgn}\bigl(R_1(i)-R_1(j)\bigr)\,
\operatorname{sgn}\bigl(R_2(i)-R_2(j)\bigr)}
{\tfrac{1}{2}\,n(n-1)}
$}
\vspace{0.6em} 

The second metric is \textbf{Position Consistency}, which quantifies the proportion of elements that occupy the same relative positions in both rankings. If $\mathrm{pos}_{R_1}(i)$ and $\mathrm{pos}_{R_2}(i)$ denote the positions of item $i$ in $R_1$ and $R_2$, respectively, then
\[
\mathrm{PC}(R_1,R_2)
=
\frac{1}{n}\,
\bigl|\{\,i \mid \mathrm{pos}_{R_1}(i)=\mathrm{pos}_{R_2}(i)\}\bigr|
\]

As shown in Tab.~\ref{tab:eval_j}, the rankings adjusted by the Judger exhibit greater alignment with human-annotated rankings. This suggests that Judger enhances the consistency of outputs with human judgments, thereby improving their quality.

\subsubsection{The Role of  Curriculum Learning}

We analyze the effects of the curriculum learning paradigm. Specifically, we compare the performance of LLaMA3-8B-Instruct with the DPO+Judger+CL method across three training stages. Stage0 represents  the base model, while Stage3 and Stage5 represent stages where the model completes the easy curriculum and the hard curriculum, respectively.

As shown in Fig.~\ref{fig:foll}, our  training paradigm progressively enhances the model's constraint following capability across various training stages. Specifically, after easy curriculum learning, the model trained in Stage3 shows superior performance compared to the base model across  L1-L3. The model's performance at L4-L5 in Stage3 is lower than Stage0. The  reason is  Stage3 has not adequately prepared for the complexity of L4-L5. The gap between these difficulty levels leads to the  performance drop. Subsequentially, when the model progresses to Stage5, after hard curriculum learning, the average performance improves significantly at the difficlut levels L4-L5.
The results on IFEval further support this conclusion. Stage0 has the lowest average performance across all indicators. After curriculum learning, there is a significant improvement in the model's performance on IFEval. Although Stage 5 performs slightly worse than Stage 3 on IFEval, Stage 5 shows a significant performance improvement on FollowBench, indicating the enhancement in the model's ability to follow soft constraints.

\section{Conclusion}
In this paper, we systematically study how to improve LLMs'  ability  to follow  soft constraints. Initially, we design a pipeline to automate the construction of datasets with high-quality outputs for soft constraint following. Based on the pipeline, we introduce a reinforcement learning training method utilizing positive and negative samples generated during the pipeline. Moreover, we propose a new training paradigm that leverages curriculum learning to enhance LLMs’ soft constraint following ability. The experiment results show that our methods enhance models’ ability to follow soft constraints effectively. 



\section{Limitations}

We discuss the limitations of our study as follows. First, 
we improve the model's ability to follow soft constraints, thereby improving its overall instruction following capability. However, even when the model's output meets all the specified constraints, it may still struggle to fully comply with complex instructions due to limitations in reasoning ability or the knowledge it masters. Additionally, while the dataset constructed  encompasses a diverse set of tasks, it may still not cover some  task types in the long tail. We consider these as  key directions for future research.

\bibliography{ref}
\clearpage

\appendix

\section{Appendix}
\label{sec:Details of Data}

\subsection{Details of Soft Constraints}
\label{sec:constraint}
We utilize GPT-4o to construct soft constraints. The three categories of soft constraints that we define are as follows:

\begin{itemize}
    \item \textbf{Soft Constraints in Content}: Content soft constraints refer to limitations associated with the data itself. These constraints govern the elements of information, the logical relationships between them, and the scope of topics that need to be covered in the response. When multiple content soft constraints are imposed, the model is required to not only generate comprehensive and coherent content but also ensure that the response aligns with the specific logical definitions and boundaries outlined by the instruction. This presents a significant challenge, as it demands both the integration of diverse elements and the maintenance of internal consistency. To address this challenge, we define the following tasks for constructing and applying content soft constraints:

    \begin{enumerate}
    \item \textbf{Inclusion of Key Elements}: The response must incorporate the key points specified in the instruction. This requires the model to effectively extract and integrate relevant information, ensuring that the essential components are included without omitting critical details.
    \item \textbf{Topic Focus}: The model must narrow the discussion to a specific subtopic, avoiding broad generalizations or irrelevant tangents. This task emphasizes the importance of maintaining focus and precision within the scope defined by the instruction.
    \item \textbf{Strict Structure}: The generated content must adhere to a predefined structure, such as being organized into coherent paragraphs, utilizing subheadings, or following a specific format. This task imposes a higher demand on the model's ability to generate well-organized and structured outputs, aligning with the required presentation structure. 
   
\end{enumerate}
    We provide the prompt template for constructing the Content Soft Constraint in Tab.~\ref{tab:p_content} and Tab.~\ref{tab:p_content1}.
   \item \textbf{Soft Constraints in Situation}: Situation soft constraints  are those related to the context within which the response is situated. These constraints require the response to be adjusted according to the context or assumptions specified in the instruction, ensuring that the content is appropriate to the given background. Such adjustments may involve factors like a particular time or location, the assumption of a specific role, or drawing conclusions based on certain premises. The response must dynamically adapt to situational changes and maintain consistency with the contextual elements. The tasks defined by these constraints can be categorized as follows:
\begin{enumerate}
    \item \textbf{Role-Playing}: The response must be framed from the perspective of a specific role or persona, ensuring alignment with the contextual expectations associated with that role.
    \item \textbf{Decision Support}: The response should provide advice or recommendations that support decision-making within a particular context.
    \item \textbf{Storytelling}: The response should construct a narrative that is situated within a defined time, location, or background, maintaining coherence with the provided contextual elements.
\end{enumerate}
 We provide the prompt template for constructing the Situation Soft Constraint in Tab.~\ref{tab:p_situation}, Tab.~\ref{tab:p_situation1}, and Tab.~\ref{tab:p_situation2}.

    \item \textbf{Soft Constraints in Style}: Style soft constraints pertain to the mode of expression, encompassing factors such as the formality or informality of tone, the level of conciseness in language, and the emotional tenor. These constraints require the response to adjust its style in accordance with the given requirements, adapting to different linguistic contexts. The following task types are defined under this category:
    \begin{enumerate}
    \item \textbf{Tone Requirement}: The generated content must adopt a specific tone, such as formal, humorous, or otherwise defined.
    \item \textbf{Language Complexity Control}: The complexity of the language used must adhere to specific standards, such as maintaining conciseness and clarity or employing academic expressions.
    \item \textbf{Emotional Expression}: The response must convey a particular emotion, such as positivity or sadness, as dictated by the context.
\end{enumerate}
 We provide the prompt template for constructing the Style Soft Constraint in Tab.~\ref{tab:p_style}.
    
\end{itemize}

\subsection{Details of Judger Reordering}
We utilize GPT-4o to reorder the outputs. We provide the prompt of Judger ranking in Tab.~\ref{tab:p_judger}  and examples of how the Judger ranks responses in Tab.~\ref{tab:cases}.
\label{sec:ajudger}

\begin{table*}
\resizebox{\linewidth}{!}{
\begin{tcolorbox}
\small
You are an Instruction Rewriting Expert. You need to rewrite \#Given Instruction\# based on \#Rewriting Requirement\#, in order to obtain a \#Rewritten Instruction\#. Basically, \#Rewritten Instruction\# should adhere to the following guidelines:\\
1. Your rewriting cannot omit the non-text parts such as the table and code in \#Given Instruction\#.\\
2. \#Rewritten Instruction\# must be reasonable and must be understood and responded to by humans.\\
3. You should try your best not to make the \#Rewritten Instruction\# become verbose, \#Rewritten Instruction\# can only add 10 to 20 words into \#Given Instruction\#.\\
\textit{\color{gray}{/* The Given Instruction */}} \\
\textbf{\{Given Instruction\}}\\
\textit{\color{gray}{/* Rewriting Requirement */}} \\
Please add one proper content constraint to the \#Given Instruction\#. The content constraints include but are not limited to:\\
1. Add a Subtask or Another Related Question.\\
2. Narrow Down the Topic: Instead of a general theme or topic, provide a more specific subset.\\
3. Set a Higher Standard: Raise the bar for what's considered acceptable or successful.\\
4. Limit Resources: Restrict the number or type of resources someone can use.\\
5. Introduce Specific Criteria: Mandate particular components or features that must be included.\\
6. Specifying Sequence: Dictate the order in which certain steps or actions should be taken.\\
Please output in JSON format with the fields 'modified\_instruction' for the modified instruction and 'added\_constraint' for the added constraint.
        
\end{tcolorbox}}
\caption{The prompt template for constructing the open-ended question answering task in Content Soft Constraint~\cite{jiang2023followbench}.}
\label{tab:p_content}
\end{table*}

\begin{table*}
\resizebox{\linewidth}{!}{
\begin{tcolorbox}
\small
You are an Instruction Rewriting Expert. You need to rewrite \#Given Instruction\# based on \#Rewriting Requirement\#, in order to obtain a \#Rewritten Instruction\#. Basically, \#Rewritten Instruction\# should adhere to the following guidelines:\\
1. Your rewriting cannot omit the non-text parts such as the table and code in \#Given Instruction\#.\\
2. \#Rewritten Instruction\# must be reasonable and must be understood and responded to by humans.\\
3. You should try your best not to make the \#Rewritten Instruction\# become verbose, \#Rewritten Instruction\# can only add 10 to 20 words into \#Given Instruction\#.\\
\textit{\color{gray}{/* The Given Instruction */}} \\
\textbf{\{Given Instruction\}}\\
\textit{\color{gray}{/* Rewriting Requirement */}} \\
Please add one proper content constraint to the \#Given Instruction\#. The content constraints include but are not limited to:\\
1. Specify Language Complexity: Determine whether the text should use simple, intermediate, or advanced language.\\
2. Control Output Length: Set limits on the text's length, such as maximum word count or number of paragraphs.\\
3. Restrict Vocabulary: Include or exclude specific words or phrases, or limit the range of vocabulary.\\
4. Mandate Structure: Require a specific format, such as headings, bullet points, or a particular narrative style.\\
Please output in JSON format with the fields 'modified\_instruction' for the modified instruction and 'added\_constraint' for the added constraint.

\end{tcolorbox}}
\caption{The prompt template for constructing the  language limitations  in Content Soft Constraint.}
\label{tab:p_content1}

\end{table*}

\begin{table*}
\resizebox{\linewidth}{!}{
\begin{tcolorbox}
\small
You are an Instruction Rewriting Expert. You need to rewrite \#Given Instruction\# based on \#Rewriting Requirement\#, in order to obtain a \#Rewritten Instruction\#. Basically, \#Rewritten Instruction\# should adhere to the following guidelines:
1. Your rewriting cannot omit the non-text parts such as the table and code in \#Given Instruction\#.\\
2. \#Rewritten Instruction\# must be reasonable and must be understood and responded to by humans.\\
3. You should try your best not to make the \#Rewritten Instruction\# become verbose, \#Rewritten Instruction\# can only add 10 to 20 words into \#Given Instruction\#.\\
\textit{\color{gray}{/* The Given Instruction */}} \\
\textbf{\{Given Instruction\}}\\
\textit{\color{gray}{/* Rewriting Requirement */}} \\
Please add one proper situation constraint to the \#Given Instruction\#. The situation constraints include but are not limited to:\\
1. Define the Context: Specify a particular situation or environment that the suggestions should be relevant to.\\
2. Introduce a Specific Problem: Focus on addressing a distinct problem or challenge that needs suggestions.\\
3. Impose Urgency: Include a time constraint or urgency for when the suggestions should be applied.\\
4. Limit Options: Restrict the scope of potential suggestions to a narrower set of choices.\\
5. Add Dependencies: Require that suggestions consider certain conditions or prerequisites.\\
6. Prioritize Outcomes: Highlight specific outcomes or goals that the suggestions should aim to achieve.\\
Please output in JSON format with the fields 'modified\_instruction' for the modified instruction and 'added\_constraint' for the added constraint.  
\end{tcolorbox}}
\caption{The prompt template for constructing the suggestion generation task in Situation Soft Constraint.}
\label{tab:p_situation}
\end{table*}

\begin{table*}
\resizebox{\linewidth}{!}{
\begin{tcolorbox}
\small
You are an Instruction Rewriting Expert. You need to rewrite \#Given Instruction\# based on \#Rewriting Requirement\#, in order to obtain a \#Rewritten Instruction\#. Basically, \#Rewritten Instruction\# should adhere to the following guidelines:
1. Your rewriting cannot omit the non-text parts such as the table and code in \#Given Instruction\#.\\
2. \#Rewritten Instruction\# must be reasonable and must be understood and responded to by humans.\\
3. You should try your best not to make the \#Rewritten Instruction\# become verbose, \#Rewritten Instruction\# can only add 10 to 20 words into \#Given Instruction\#.\\
\textit{\color{gray}{/* The Given Instruction */}} \\
\textbf{\{Given Instruction\}}\\
\textit{\color{gray}{/* Rewriting Requirement */}} \\
Please add one proper situation constraint to the \#Given Instruction\#. The situation constraints include but are not limited to:\\
1. Specify a Role: Clearly define the role or persona to be taken on during the role-play.\\
2. Define the Setting: Outline the environment or context in which the role-play should occur.\\
3. Add Conflict or Challenge: Introduce a specific problem, conflict, or challenge that must be addressed within the role-play.\\
4. Limit the Actions: Restrict the types or number of actions that can be taken during the role-play.\\
5. Set Specific Goals: Define clear objectives that the role-player must achieve.\\
6. Introduce Time Constraints: Impose a time limit for the role-play to unfold or for certain actions to be completed.\\ 
Please output in JSON format with the fields 'modified\_instruction' for the modified instruction and 'added\_constraint' for the added constraint.  
\end{tcolorbox}}
\caption{The prompt template for constructing the role-playing task in Situation Soft Constraint.}
\label{tab:p_situation1}
\end{table*}

\begin{table*}
\resizebox{\linewidth}{!}{
\begin{tcolorbox}
\small
You are an Instruction Rewriting Expert. You need to rewrite \#Given Instruction\# based on \#Rewriting Requirement\#, in order to obtain a \#Rewritten Instruction\#. Basically, \#Rewritten Instruction\# should adhere to the following guidelines:
1. Your rewriting cannot omit the non-text parts such as the table and code in \#Given Instruction\#.\\
2. \#Rewritten Instruction\# must be reasonable and must be understood and responded to by humans.\\
3. You should try your best not to make the \#Rewritten Instruction\# become verbose, \#Rewritten Instruction\# can only add 10 to 20 words into \#Given Instruction\#.\\
\textit{\color{gray}{/* The Given Instruction */}} \\
\textbf{\{Given Instruction\}}\\
\textit{\color{gray}{/* Rewriting Requirement */}} \\
Please add one proper situation constraint to the \#Given Instruction\#. The situation constraints include but are not limited to:\\
1. Define Character Archetypes: Specify certain archetypes or roles characters should fulfill, such as a hero, mentor, or antagonist.\\
2. Include Specific Plot Points: Mandate the inclusion of certain events or plot twists that must occur.\\
3. Moral Dilemmas: Introduce a scenario where the characters must make a tough decision that involves competing ethical principles or risks.\\
Please output in JSON format with the fields 'modified\_instruction' for the modified instruction and 'added\_constraint' for the added constraint.  
\end{tcolorbox}}
\caption{The prompt template for constructing the story generation task in Situation Soft Constraint.}
\label{tab:p_situation2}
\end{table*}

\begin{table*}
\resizebox{\linewidth}{!}{
\begin{tcolorbox}
\small
You are an Instruction Rewriting Expert. You need to rewrite \#Given Instruction\# based on \#Rewriting Requirement\#, in order to obtain a \#Rewritten Instruction\#. Basically, \#Rewritten Instruction\# should adhere to the following guidelines:
1. Your rewriting cannot omit the non-text parts such as the table and code in \#Given Instruction\#.
2. \#Rewritten Instruction\# must be reasonable and must be understood and responded to by humans.
3. You should try your best not to make the \#Rewritten Instruction\# become verbose, \#Rewritten Instruction\# can only add 10 to 20 words into \#Given Instruction\#.\\
\textit{\color{gray}{/* The Given Instruction */}} \\
\textbf{\{Given Instruction\}}\\
\textit{\color{gray}{/* Rewriting Requirement */}} \\
Please add one proper style constraint to the \#Given Instruction\#. The style constraints include but are not limited to:\\
1. Tone and Emotion: Specify the desired emotional tone for the response.\\
2. Writing Style: Ask the AI to mimic a specific author's writing style.\\
3. Contradiction: Ask the AI to provide a response that contradicts the previous statement or take a stance opposite to its prior response.\\
4. Ambiguity: Instruct the AI to create responses with intentional ambiguity or double meanings.\\
5. Humor or Satire: Request that the response be humorous or satirical, requiring the AI to generate jokes or witty remarks.\\
       
\end{tcolorbox}}
\caption{The prompt template for constructing the Style Soft Constraint~\cite{jiang2023followbench}.}
\label{tab:p_style}
\end{table*}
\begin{table*}
\resizebox{\linewidth}{!}{
\begin{tcolorbox}
\small
You are a helpful assistant who reviews a debate between two other assistants in
evaluating the quality of the outputs for a given instruction.The two assistants, Assistant (a) and Assistant (b), are given an instruction.
Output (a) and Output (b) are generated
by two different AI chatbots respectively. 
Assistant (a) and Assistant (b) have conflicting evaluations. Your goal is to review
their evaluations and give your final decision on which output is better.
Here are some rules of the evaluation: \\
(1) You should prioritize evaluating whether the output honestly/precisely/closely
executes the instruction, then consider its helpfulness, accuracy, level of detail,
harmlessness, etc. \\
(2) Outputs should NOT contain more/less than what the instruction asks for, as
such outputs do NOT precisely execute the instruction. \\
(3) You should avoid any potential bias and your judgment should be as objective
as possible. For example, the order in which the outputs were presented should
NOT affect your judgment, as Output (a) and Output (b) are **equally likely** to
be the better.\\
Output your final verdict by strictly following this format: 
"[[A]]" if Output (a) is better, "[[B]]"
if  Output (b) is better, and "[[C]]" for a tie.\\
\textit{\color{gray}{/* Given instruction */}} \\ 
\textbf{\{question\}} \\
\textit{\color{gray}{/* The Start of Output (a) */}} \\
\textbf{\{answer of assistant a\}} \\
\textit{\color{gray}{/* The End of Output (a) */}} \\
\textit{\color{gray}{/* The Start of Output (b) */}} \\
\textbf{\{answer of assistant b\}} \\
\textit{\color{gray}{/* The End of Output (b) */}} \\
\end{tcolorbox}}

\caption{The prompt template for Judger to reorder the responses~\cite{zheng2023judging}}.
\label{tab:p_judger}
\end{table*}

\subsection{Implementation Details}
We train LLaMA-3-8B-Instruct, Mistral-7B-Instruct-v0.3, and LLaMA2-13B-Chat-HF  using LLaMA-Factory~\cite{zheng2024llamafactory} on 4 NVIDIA A100 80GB GPUs, applying LoRA~\cite{hu2021lora} for efficient training. The lora target is set to all, with all models training for 3 epochs. The per device train batch size is set to 1, and gradient accumulation steps is set to 8. The warm-up ratio is set to 0.1.
For SFT, LLaMA-3-8B-Instruct has a learning rate of 1.0e-4, Mistral-7B-Instruct-v0.3 uses 5.0e-7, and LLaMA2-13B-Chat-HF uses 1.0e-7. For DPO, the learning rate is 5.0e-6 with a beta value of 0.1. We apply cosine learning rate scheduler. For the benchmark evaluation, we adopt the online serving inference approach using vLLM.




\end{document}